\documentclass{article}
\usepackage[final]{neurips_2021}
\usepackage[utf8]{inputenc} % allow utf-8 input
\usepackage[T1]{fontenc}    % use 8-bit T1 fonts
\usepackage{amsmath,amssymb,amsfonts}
\usepackage{graphicx}
\usepackage{textcomp}
\usepackage{xcolor}
\usepackage{microtype}      % microtypography
\usepackage{caption}
\usepackage{subcaption}
\usepackage{graphicx}
\usepackage{amsthm}
\usepackage{algorithm}
\usepackage{soul}
\usepackage{url}
\usepackage{hyperref}
\usepackage{booktabs}
\usepackage{nicefrac}       % compact symbols for 1/2, etc.
\usepackage[noend]{algpseudocode}
\newtheorem{cor}{Corollary}

\newtheorem{prop}{Proposition}
\newtheorem{lem}{Lemma}
\newtheorem{ass}{Assumption}
\newtheorem{thm}{Theorem}

\newcommand{\BEA}{\begin{eqnarray}}
\newcommand{\EEA}{\end{eqnarray}}
\newcommand{\beq}{\begin{equation}}
\newcommand{\eeq}{\end{equation}}

\newcommand{\argmin}{\mathop{\mathrm{argmin}}}
\newcommand{\prox}{\operatorname{prox}}

\def \sign{{sign}}
\def \eye{\mathrm{I}}
\def \zer{\mathbf{0}}
\def \prb{\mathbb{E}}
\def \RR{\mathbb{R}} 
\def \ww{\mathbf{w}} 
\def \wg{\tilde{\mathbf{w}}} 
\def \W{\mathbf{W}} 
 
\def \zz{\mathbf{z}} 
\def \V{{\boldsymbol{\Theta}}} 
\def \vv{{\boldsymbol{\theta}}} 
\def \A {\mathcal{A}} %%% aggregation function
\def \K {{N}} 
\def \S {\mathcal{S}} 
\def \N {\mathcal{N}} 

\makeatletter
\newcommand{\printfnsymbol}[1]{%
  \textsuperscript{\@fnsymbol{#1}}%
}
\makeatother

\title{Robustness and Personalization in Federated Learning: A Unified Approach via Regularization
}
\author{%
  Achintya Kundu\printfnsymbol{1} \\
  IBM Research, Singapore \\
  \texttt{achikundu@gmail.com} \\
  \And
  Pengqian Yu\thanks{Denotes equal contribution.} \\
  IBM Research, Singapore \\
  \texttt{yupengqian1989@gmail.com} \\
  \AND
  Laura Wynter \\
  IBM Research, Singapore \\
  \texttt{lwynter@sg.ibm.com} \\
  \And
  Shiau Hong Lim \\
  IBM Research, Singapore \\
  \texttt{shonglim@sg.ibm.com} \\
}

\begin{document}
\maketitle
\begin{abstract}
 We present a class of methods for robust, personalized federated learning,  called Fed+, that unifies many federated learning algorithms. The principal advantage of this class of methods is to better accommodate the real-world characteristics found in federated training, such as the lack of IID data across parties, the need for robustness to outliers or stragglers, and the requirement to perform well on party-specific datasets. We achieve this through a  problem formulation that allows the central server to employ robust ways of aggregating the local models while keeping the structure of local computation intact. Without making any statistical assumption on the degree of heterogeneity of local data across parties, we provide convergence guarantees for Fed+ for convex and non-convex loss functions under different (robust) aggregation methods. The Fed+ theory is also equipped to handle heterogeneous computing environments including stragglers without additional assumptions; specifically, the convergence results cover the general setting where the number of local update steps across parties can vary. We demonstrate the benefits of Fed+  through extensive experiments across standard benchmark datasets.
\end{abstract}

%%%%%%%%%%%%%%%%%%%%%%%%%%%%%%%%%%%%%%%%%%%%%%%%%%%%%%%%%%%%

%%%%=================================================
\section{Introduction}
%%%%=================================================

Federated learning (FL) is a technique for training machine learning models without sharing data, introduced by \citet{mcmahan2017communication} and \citet{konevcny2015federated, konevcny2016federated}, and steadily gaining momentum. Federated learning involves a possibly varying set of parties participating in a parallel training process through a centralized aggregator that has access only to the parties' model parameters or gradients but not to the data itself. Compared to parallel stochastic gradient descent (SGD), federated learning aims at minimizing communication by parties performing a number of iterations locally before sending parameters to the aggregator. Federations tend to be diverse, leading to non-IID data across parties, and often include parties whose data can be considered to be outliers with respect to the others. Most algorithms, however, can trigger a failure of the training process itself when parties are too heterogeneous in precisely the settings where federated learning could have the greatest benefit. Personalization of federated model training, when judiciously performed, is one means of avoiding such training failure. In addition, personalization of federated training allows for greater accuracy on the data that matters most to each party. The majority of federated learning fusion algorithms are designed to produce a common solution for all parties. However, this is seldom the setting that motivates the use of federated learning. As also noted by \citet{3personalized}, an application (e.g. of sentence completion) for a  user should be optimized for that user's needs and not be identical across all users.

We propose Fed+ (pronounced as FedPlus) to address the issues of avoiding training failure, increasing robustness to outliers and stragglers, and improving performance on the applications of interest where party-level data distributions need not be similar across parties. Fed+ unifies many algorithms and offers provably-convergent personalization and robustness; this is achieved through a problem formulation that allows the central server to employ robust ways of aggregating the local models while keeping the structure of local computation intact. 

Fed+ does not make explicit assumptions on the distributions of the local data, which are assumed private to each party. Instead, we assume a global shared parameter space with locally computed loss functions.    Like some personalized methods, Fed+  allows for data heterogeneity by relaxing the requirement that the parties must reach a full consensus. The  Fed+ theory is equipped to handle heterogeneous computing environments, including stragglers, without making additional assumptions; specifically, the convergence results cover the general setting where the number of local update steps across parties can vary. 

To evaluate the performance of a  federated learning aggregation method, it is important to assess it on the types of datasets on which it would be ultimately used. On the one hand, parties involved in federated learning training wish to enjoy improved accuracy on data from their own data populations. In addition,  parties involved in federated model training also aim to train models that will transfer well. As such, it is crucial to evaluate algorithms on test sets that include some data from outside the party-specific training data. We thus illustrate the benefits of Fed+  on the synthetic dataset created for FedProx by \citep{li2020federated} as well as on the LEAF datasets of \citet{caldas2018leaf} to represent the party-specific dataset scenario. We also construct personalized FL datasets on a synthetic regression problem and from the well-known MNIST dataset to provide an assessment of transfer quality within a party-specific setting.

The contributions of this work are (i) the definition of a unified framework for robust, personalized federated learning, called Fed+; (ii) a convergence theory that covers the most important variants of the Fed+ algorithm, including convex and nonconvex loss functions, robust aggregation and stragglers; and (iii) a comprehensive set of numerical experiments on party-specific datasets with and without with transfer requirements,  thus illustrating the benefit of Fed+ with respect to other federated learning algorithms, personalized and non-personalized.

%%%%====================================================================================
\section{Related Work}
%%%%====================================================================================
\citet{ConvFedAvg} showed that FedAvg defined by \citet{mcmahan2017communication} can converge to a point that is not a solution to the original problem and proposed to add a  decreasing learning rate; with that, they provide a theoretical convergence guarantee, even when the data is not IID, but the resulting algorithm is slow to converge. To handle non-IID data, \citet{li2020federated} introduced a  regularization term in their FedProx algorithm.  \citet{ConvFedAvg,2019SCAFFOLD} seek to explain the non-convergence of  FedAvg while proposing new algorithms.
\citet{Pathak2020FedSplit,Charles2020Local,Malinovsky2020} propose FedSplit and LocalUpdate, and Local Fixed Point, resp., and obtain tight bounds on the number of communication rounds required to achieve an $\epsilon$ accuracy. 
However, these algorithms all require the convergence of all parties to a  common model.
Others have sought to increase robustness to corrupted updates and outliers.  \citet{pillutla2019robust} proposed Robust Federated Aggregation (RFA) by replacing the weighted arithmetic mean aggregation with an approximate geometric median.  \citet{pmlr-v80-yin18a} proposed a Byzantine-robust distributed statistical learning algorithm based on the coordinate-wise median. Both  RFA \citep{pillutla2019robust} and coordinate-wise median \citep{pmlr-v80-yin18a}
involve training a single global model, and neither is robust to non-IID data, leading in some cases to failure of the learning process. 

Several recent works advocate, as we do, for a fully personalized approach whereby each client trains a local model while contributing to a global model. \citet{3personalized} proposed clustering parties and solving an aggregate model within each cluster. While this would likely eliminate the training failure we observe in practice, it adds considerable overhead. \citet{KAUSTLocalGlobal} proposed a local-global mixture method focused on reducing communication overhead for the smooth convex setting. \citet{deng2020adaptive} proposed a method  similar to our FedAvg+. \citet{moreau} proposed a  procedure for mean aggregation where each party optimizes its local loss and a (local version of) the global parameters. \citet{unified_personalized} provided a unification of mean personalized aggregation for smooth and convex loss functions. \citet{ditto} proposed a bilevel programming framework that alternates between solving for the mean aggregate solution and the local solutions. The overall problem, however, is non-convex, even when parties have convex loss functions, and could be solved in two separate phases. \citet{localcommonsoln} suggested personalizing the mean aggregate solution as a set of weighted average aggregate solutions.
The most important difference between Fed+ and the above methods is that only Fed+ allows for robust aggregation, both in the definition of the algorithm and in the convergence theory, handling the resulting non-smooth optimization problem.

%%%%%%%%%%%%%%%%%%%%%%%%%%%%%%%%%%%%%%%%%%%%%%%%%%%%%%%%%%%%
\section{Illustration of Training Failure in Federated Learning }
\label{sec:training_failure}

\begin{figure}
	\centering
	\includegraphics[width=0.8\linewidth]{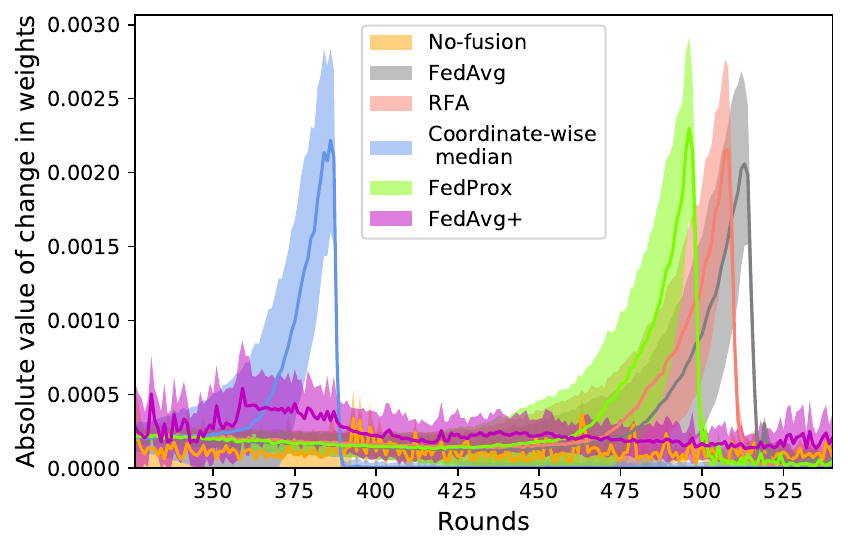}
	\caption{Change in  weights before and after each aggregation round. Only Fed+ and local SGD without FL ("no fusion")  have no large spikes.}
	\label{fig:weights_change}
\end{figure}

Here, we illustrate the training failure that can occur in real-world federated learning settings on a federated reinforcement learning-based financial portfolio management problem. The key observation, see Figure \ref{fig:weights_change}, is that replacing the local party models with a  common, aggregate model at each round can lead to large spikes in model changes,  triggering training failure for the federation as a whole. The figure shows the mean and standard deviation of the change in neural network parameter values before and after a federated learning aggregation step. FedAvg, RFA using the geometric median,  coordinate-wise median, and FedProx are shown, as well as the no fusion case where each party trains independently on its own data, and the FedAvg+ version of  Fed+. All the standard FL methods cause large spikes in the parameter change that do not occur without federated learning or with Fed+. 

\begin{figure}
	\centering
	\includegraphics[width=0.8\linewidth]{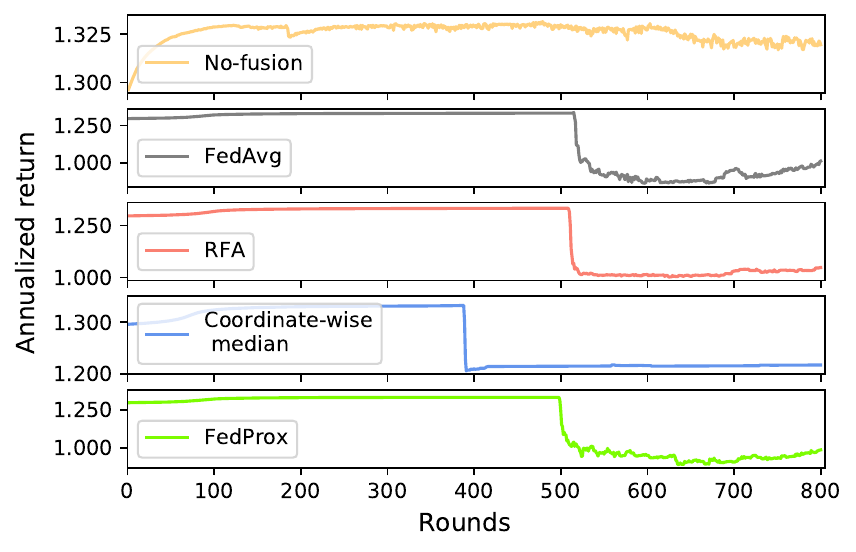}
	\caption{Illustration of  training collapse experienced using all standard methods except local SGD without FL (``no fusion"). }
	\label{fig:training_2}
\end{figure}

Such dramatic model change can lead to a collapse of the training process. The large spikes coincide precisely with training collapse, as shown in Figure \ref{fig:training_2} (bottom four figures).
Note that this example \emph{does not involve adversarial parties or party failure}, as evident from the fact that single-party training (top curve) does not suffer failure. Rather, it shows a real-world problem where parties' data are not drawn IID  from a single dataset. It is conceivable that federated training failure may be a   common occurrence in practice when forcing convergence to a common solution across parties. 

\begin{figure}
	\centering
	\includegraphics[width=0.8\linewidth]{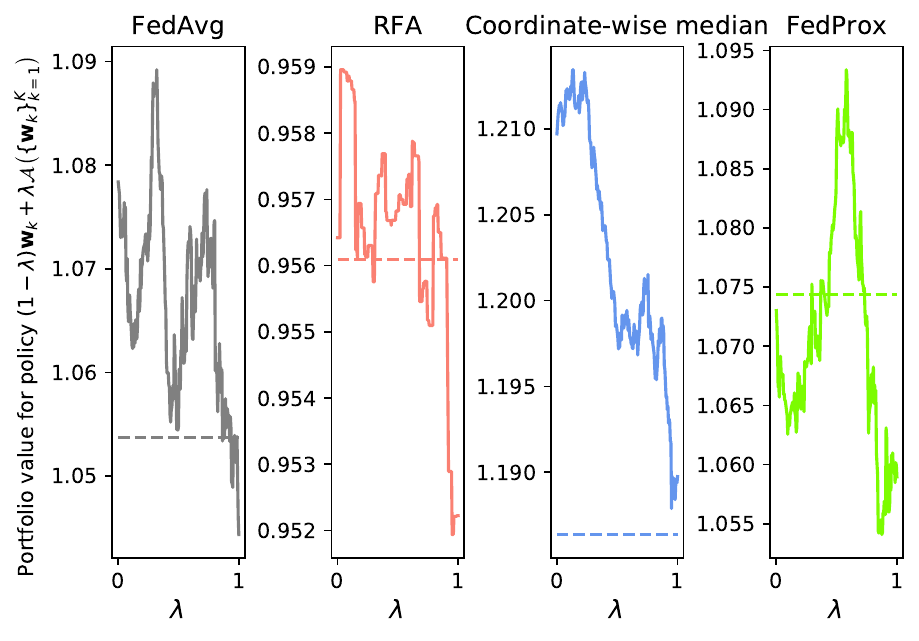}
	\caption{Before and after  federated model aggregation, along $\lambda \in[0,1]$ between the local ($\lambda=0$) and global ($\lambda=1$) solutions.}
	\label{fig:lambda}
\end{figure}

A deeper understanding of the training failure can be gleaned from Figure \ref{fig:lambda}, which shows what occurs before and after an aggregation step and motivates the Fed+ approach. 
A local party update occurs in each subplot on the left side, at  $\lambda=0$. Values of $\lambda \in [0,1]$ correspond to moving towards, but not reaching, the common, aggregate model. A right-hand side lower than the left-hand side means that \textit{a full step towards averaging (or using the median for)} all parties, i.e., $\lambda=1$, degrades local performance. Dashed lines represent the aggregated model in the previous round. Observe that local updates improve the performance from the previous aggregation indicated by the dashed lines. However,  performance degrades after the subsequent aggregation, corresponding to the right-hand side of each subplot, where $\lambda=1$. In fact, for FedAvg, RFA, and FedProx, the performance of the subsequent aggregation is worse than the previous value (dashed line).

%%%%====================================================================================

%%%%%%%%%%%%%%%%%%%%%%%%%%%%%%%%%%
\section{The Fed+ Framework}
%%%%%%%%%%%%%%%%%%%%%%%%%%%%%%%%%%

 We design the Fed+ framework to handle real-world federated learning settings better, including non-IID data across parties,  parties having outlier data with respect to other parties, stragglers, in that updates are transmitted late, and an implicit requirement for the final trained model(s) to perform well on both each party's own datasets as well as datasets whose distributions differ from the party's training data. To accomplish these goals, Fed+ takes a robust, personalized approach to federated learning and, importantly, does not require all parties to converge to a single central point. Fed+ thus requires generalizing the objective of the federated learning training process, as follows.

\subsection{Problem Formulation} Consider a federation of $\K$ parties with local loss functions $f_k:\RR^d \to \RR, k=1,2,\ldots, \K$. The original FedAvg formulation \citep{mcmahan2017communication} involves  training a central model $\wg \in \RR^d$ by minimizing the average local loss over the $\K$ parties: 
\BEA
	\min_{\W, \,\wg}  ~\Big[\,F(\W) ~:=~ \frac{1}{\K} \sum_{k=1}^{\K} f_k(\ww_k)  \,\Big]  ~~~ \textrm{subject to} \quad  \ww_k = \wg,\, \, k=1,\ldots,\K, 
	\label{eq:FA}
\EEA
where we use the notation $\W := (\ww_1,\ww_2,\dots, \ww_\K) \in \RR^{d \times \K}$ with $\ww_k \in \RR^d$ denoting the local model of party $k$. 

Fed+ proposes learning personalized models of the form $\ww_k = \wg+\vv_k$, where the personalized component $\vv_k$ is regularized through a choice of convex function $\Psi:\RR^d \to [0,\infty]$, that is: 
\BEA
	 \min_{\W,\V, \,\wg}  ~\frac{1}{\K}  \sum_{k=1}^{\K} \Big[\, f_k (\ww_k) \,+ \, \Psi(\vv_k) \,\Big]  ~~~
	\textrm{subject to} \quad  \ww_k = \wg+\vv_k, \,\, k=1,\ldots,\K, 
	\label{eq:FP}
\EEA
where $\V := (\vv_1,\vv_2,\dots, \vv_\K) \in \RR^{d \times \K}$. Note that \eqref{eq:FA} is a special case of \eqref{eq:FP} when we set $\Psi(\vv) = 0$ if $\vv = \zer$ and $+\infty$ otherwise. In this work, we explore robust regularization functions like $\|\cdot\|_1$ and $\| \cdot \|_2$ as well as the usual squared Euclidean norm. Now, in place of the hard equality constraints in \eqref{eq:FA}, Fed+ takes a  penalization-based approach, resulting in  the following objective for the overall Fed+ federated training process:
\BEA
     [\mbox{\emph{ Fed+~Optimization~Problem }}]  \hspace{4cm} \nonumber \\
     \displaystyle \min_{\W, \V, \wg} ~H_{\sigma}(\W,\V,\wg) ~:=~ \frac{1}{\K}\sum_{k=1}^{\K}  \left[\, f_k(\ww_k) \,+ \, \Psi(\vv_k) \,+\, \frac{\sigma}{2}  \| \ww_k - (\wg + \vv_k)\|_2^2 \, \right],
  \label{eq:FPz}
\EEA
where $\sigma >0$ is a user-chosen penalty constant. 

\subsection{Robust Aggregation} 
Let $\A$ denote an aggregation function that outputs a central aggregate $\wg \in \RR^d$  of $\ww_1,\dots, \ww_\K$. That is,  the global model $\wg$ is computed by aggregating the current local models $\{\ww_1,\ldots,\ww_\K \}$. The geometric median and coordinate-wise median aggregation functions are defined, respectively, by 
\BEA
\mathtt{Geometric Median}\left( \ww_1, \, \cdots, \,\ww_\K \right) ~ := ~\displaystyle \argmin_{ \ww \in \RR^d }~ \frac{1}{\K} \sum_{k=1}^{\K} \|\ww_k -\ww\|_2.
\EEA
\BEA
\mathtt{Coordinatewise Median}\left( \ww_1, \, \cdots, \,\ww_\K \right)  ~:= ~\displaystyle \argmin_{ \ww \in \RR^d }~ \frac{1}{\K} \sum_{k=1}^{\K} \|\ww_k -\ww\|_1.
\label{eq:CoMed}
\EEA
%\BEA \begin{array}{l} \mathtt{Mean}\left(\{ \ww_1, \, \cdots, \,\ww_\K \}\right)  :=  \displaystyle \argmin_{ \ww \in \RR^d }~ \frac{1}{\K} \sum_{k=1}^{\K} \|\ww_k -\ww\|_2^2. \end{array} \nonumber \EEA $$\|\ww-\ww'\|_2 \; \mbox{and}\;  \|\ww-\ww'\|_1,$$ and  mean  aggregation by $\|\ww-\ww'\|_2^2$.  This leads  to the following definitions:
Note that computing robust aggregation functions such as geometric median and coordinate-wise median involve non-smooth optimization. Fed+ unifies smooth and non-smooth aggregation through smoothing with parameter $\delta>0$ by employing $\hat{\Psi}_{\delta}$, a $(1/\delta)$-smoothed approximation of $\Psi$, known as the Moreau envelope of $\Psi$:
\BEA
    \hat{\Psi}_{\delta}(\ww) ~:=~ \displaystyle \min_{ \vv \in \RR^d}\left[\,\Psi(\vv) + \frac{1}{2\delta}\|\ww - \vv\|_2^2\,\right],
    \label{eq:Bphi}
\EEA
where the minimizer in \eqref{eq:Bphi} is called the proximal operator of $\Psi$ and is denoted by $\prox^{\delta}_{\Psi}(\ww)$. The Fed+ aggregation function $\A$ is then defined in terms of the regularization function $\Psi$ as follows:
\BEA
    \A(\W) ~:=~  \displaystyle \argmin_{ \ww \in \RR^d }~ \frac{1}{\K} \sum_{k=1}^{\K} \hat{\Psi}_{\frac{1}{\sigma}}(\ww_k -\ww). \label{eq:aggregation_def}
\EEA
Therefore, by choosing $\Psi$ to be the scaled $\ell_2$ norm, $\sigma \delta \|\cdot\|_2$ to be precise, we obtain a $\delta$-approximation of the geometric median aggregation as used in \citet{pillutla2019robust}. Similarly, setting $\Psi$ to the scaled $\ell_1$ norm ($\sigma \delta \|\cdot\|_1$) gives a $\delta$-approximation of the coordinate-wise median aggregation. The usual mean aggregation is naturally recovered in both of these cases:  $\Psi(\vv) = \frac{\sigma \delta}{2}\|\vv\|_2^2$, and  $\Psi(\vv) = 0$ if $\vv = \zer$ and $+\infty$ otherwise.

\subsection{Personalization at the Local Parties}
The personalized federated setting involves  each active party $k$  solving its own model that includes  a party-specific loss and the aggregate parameter value $\tilde{w}$. At every round, each party $k$ runs $E_k$ iterations of the following two-step update rule, with learning rate $\eta >0$, though in practice, the exact gradient $\nabla f_k(\ww_k)$ is replaced by  an unbiased random estimate. Specifically,
\BEA \begin{array}{l}
	\ww_k ~\gets~  \ww_k - \eta \nabla f_k (\ww_k) ,\\
	\ww_k ~\gets~  \kappa \,\ww_k \,+\, (1-\kappa)\,[\wg^{t}+\vv_k^t],
	\end{array}
	\label{eq:xk_update}
\EEA
where the constant $\kappa:= \frac{1}{1+\sigma\eta} \in (0,1]$  controls the degree of regularization used for training the local model and personalization occurs via the  party-specific regularization term $\vv_k$. A natural choice for $\vv_k \in \RR^d$ is to use a robust function of the difference between the current local and global model. That is, Fed+ proposes setting $\vv_k$ by minimizing \eqref{eq:FPz} w.r.t. $\vv_k$  keeping $\ww_k$ and $\wg$ fixed, leading to the  closed form update:
\BEA
    \vv_k ~\gets ~ \prox^{\frac{1}{\sigma}}_{\Psi} ( \ww_k - \wg).
    \label{eq:zk_rp_def}
\EEA
The party-specific, personalized gradient update of Fed+ is summarized in Proposition \ref{prop:F_grad_step} below.

 \begin{prop}\label{prop:F_grad_step}
The local, personalized update  in the Fed+ algorithm is a gradient descent iteration with learning rate $\eta\kappa$ where $\kappa:= \frac{1}{1+\sigma\eta}$, applied to the following sub-problem: 
\BEA
	\min_{\ww_k \in \RR^d} ~ \left[ \, F_k(\ww_k;\vv_k,\wg) ~:=~ f_k(\ww_k) + \frac{\sigma}{2}   \|\ww_k - (\wg+\vv_k) \|_2^2  \,\right] , \label{eq:Fk_def}
\EEA 
where $\vv_k$ \& $\wg$ are kept fixed.
\end{prop}

\begin{algorithm}[hbt!]
	\caption{Fed+: parties $k=1\dots\K$;
	number of federated training rounds $T$, 
	number of active parties per round $K$, 
	number of local iterations per round at party $k$, $E_k$;  
	learning rate $\eta >0$; 
	penalty constant $\sigma > 0$; 
	regularization function $\Psi: \RR^d \to [0,\infty]$; and 
	 local initialization parameter, $\lambda = 0$.} 
	\label{alg:FP}
	\begin{algorithmic}[1]
		\Statex \textbf{Initialization:}
		\State Each party $k$ initializes local model $\ww_k^{0}$ to the Aggregator's global model $\wg^0$.
		\Statex \textbf{Aggregator:}		
		\For{round $t=0,\dots,T-1$}
		\State Sample a subset $\S^t$ of size $K$ from $\{1,\dots,\K\}$. % uniformly at random  .
		\State Send the global model $\wg^{t}$ to each party $k\in \S^t$.		
		\For{each party $k\in \S^t$ \textbf{in parallel}} 
		\State $\ww_k^{t+1} \gets \text{Local-Solve}(k,\,t,\,\wg^{t},\,\ww_k^t)$. 
		\State Party sends $\ww_k^{t+1}$ to Aggregator. 
		\EndFor
	    \For{each party  $k \notin \S^t$}
	    \State Set $\ww_k^{t+1} \gets \ww_k^{t}$.
		\EndFor
		\State Compute the global model by aggregating the local \\ \hspace{1cm} models:~  $\wg^{t+1} ~\gets~ \A(\W^{t+1})$, ~ where \\
		 \hspace{1cm}$\A(\W^{t+1}) := \displaystyle  \argmin_{\wg} \frac{1}{\K} \sum_{k=1}^{\K} \hat{\Psi}_{\frac{1}{\sigma}}(\ww_k^{t+1}-\wg)
		$. \label{alg_aggr_step}
		\EndFor
		\Statex \textbf{ Local-Solve }$(k, \,t, \,\wg^{t}\,\ww_k^t)$:     \quad // Run on each $k \in \S^t$
		\State Compute the personalized component for regularization: \\
		\hspace{3cm} $ \vv_k^{t} \gets \prox^{\frac{1}{\sigma}}_{\Psi}(\ww_k^{t}-\wg^{t}).  $ \label{alg_personal_model}
		\State Initialize the local model: 
        $ \ww_{k}^{t+1} ~\gets~ %\ww_k^{t}.
		    (1-\lambda)\ww_k^{t} + \lambda \wg^{t}$. 	\label{alg_local_init}
		\For{$i=0,1,\dots,(E_k-1)$}
		\BEA \begin{array}{l}
		%\ww_{k,i+\frac{1}{2}}^{t+1} \gets  \ww_{k,i}^{t+1} - \eta \nabla f_k (\ww_{k,i}^t), \\
		%\ww_{k,i+1}^{t+1} \gets  \displaystyle \left( \frac{1}{1+\eta\sigma} \right) \ww_{k,i+\frac{1}{2}}^{t+1} + \left( \frac{\eta\sigma}{1+\eta\sigma} \right)[\wg^t + \vv_k^{t}]. \\
		\ww_{k}^{t+1} ~\gets ~ \kappa  \displaystyle \left[ \ww_{k}^{t+1} - \eta \nabla f_k (\ww_{k}^{t+1}) \right] \,+ \, (1-\kappa)\left[  \wg^t + \vv_k^{t} \right], ~\kappa \,:=\,  \frac{1}{1+\eta\sigma}.
		% \\ \mbox{where}~ \kappa \,:=\,  \frac{1}{1+\eta\sigma}.
		\end{array}  \label{eq:alg_local_update}
		\EEA 
		\EndFor		
		\State Return  $\ww_{k}^{t+1}$. \\
	\end{algorithmic}
\end{algorithm}

\subsection{The Fed+ Algorithm}  
Fed+ is defined in Algorithm \ref{alg:FP} to solve \eqref{eq:FP} with  \eqref{eq:Bphi}. Fed+ is designed to allow for robust aggregation functions $\A$, where local copies of shared parameters are aggregated. Fed+ does not require all parties to agree on a single common model. We argue that this offers the benefits of the federation without the pitfall of training failure that can occur in real-world implementations of federated learning. So as to unify important special cases, Algorithm \ref{alg:FP} introduces a number of parameters: $\lambda \in [0,1]$, $\sigma > 0$, and $\Psi:\RR^d \to [0,\infty]$. A main difference between Fed+ and other federated algorithms is that in other FL approaches, parties set the aggregate central model (which corresponds to setting $\lambda = 1$) as their starting point for their local updates. On the other hand, Fed+ advocates initializing each local model at each round with its own last value from the previous round, i.e., $\lambda = 0$. This mitigates the dramatic changes in local models that can occur in federated learning.

\subsubsection{Proposed Variants of Fed+}
We introduce three variants of interest of Fed+, unified through their choice of function $\Psi$. Furthermore, using Fed+, the variants can be combined in a hybridization approach described below. The proximal regularization constant $\sigma >0$ is a tunable hyper-parameter; we recommend setting it to a value that results in $\kappa \in [0.9,0.999]$. We set the smoothing approximation constant $\delta$ to $0.1$ and the initialization parameter $\lambda$ to $0$ unless mentioned otherwise.

{\bf FedAvg+}: A mean-aggregation based method with better training performance than FedAvg via personalization. Choose $\Psi(\ww) = \frac{\sigma \delta}{2}\|\ww\|_2^2$. This choice of $\Psi$ leads to the mean as the aggregation function $\A$ in Fed+ (see eqn. \eqref{eq:aggregation_def}), i.e., $\wg^{t+1} = \frac{1}{S}\sum_{k \in \S^t } \ww_k^{t+1}$, and the personalization component $\vv_k^t$ becomes a scaled version of the difference between the $k$-th party's current local model $\ww_k^t$ and the aggregated global model $\wg^t$: ~$\vv_k^t ~=~ [1+\delta]^{-1}(\ww_k^{t}-\wg^{t}).$

{\bf FedGeoMed+}: A robust aggregation based method that offers stability in training in the presence of outliers/adversaries. 
	Set $\Psi(\ww) =  \sigma \delta \|\ww\|_2$. In this case,  aggregation function $\A$ is a $\delta$-approximation of the Geometric Median, and the personalization component $\vv_k^t$ is given by
	\BEA
	\vv_k^t = ~ \max \left\{0, \, 1-(\delta /\|\ww_k^{t}-\wg^t\|_2) \right\}(\ww_k^{t}-\wg^{t}). \nonumber
	\EEA
	Clearly, the personalization component $\vv_k^t = \zer$ when the local model $\ww_k^t$ is close to the global model $\wg^t$, to be precise, when $\|\ww_k^{t}-\wg^t\|_2 \le \delta$. To compute the global model $\wg^{t+1}$ from $\{\ww_k^{t+1} \,:\, k\in \S^t \}$, the aggregator runs the following two step iterative procedure initialized with $\wg = \ww_{mean} := \mathtt{Mean}\{ \ww_k^{t+1}\, : \,k \in \S^t\}$ until $\wg$ converges: 
	\BEA
	\begin{array}{l}
	\vv_k  \gets  \max \left\{ 0 ,\, 1 - \frac{\delta}{\| \ww_k^{t+1} - \wg \|_2} \right\} (\ww_k^{t+1} -\wg), \,\forall \,k \in \S^t, \nonumber \\
	\wg   \gets  \ww_{mean} -\mathtt{Mean}\{ \vv_k \,:\, k \in \S^t \}.
	\end{array}\nonumber
	\EEA	

 {\bf FedCoMed+}: ~ This version offers the benefit of robust aggregation via the median with added flexibility in allowing each coordinate of the model vector to be computed independently. This is achieved through the following choice of robust regularization: $\Psi(\ww) = \sigma \delta \|\ww\|_1$. Here, the aggregation function $\A$ is a $\delta$-approximation of the Coordinate-wise Median, and the personalization component $\vv_k^t$takes the following form: 
	\BEA\begin{array}{ll}
		\vv_k^{t} & = \mathtt{Soft\_Thresholding}(\ww_k^t - \wg^t, \,\delta ) \\
		& :=\max\{ \zer , \, [\ww_k^t - \wg^t]-\delta \sign(\ww_k^t- \wg^t) \},
	\end{array}\label{eq:soft_thresholding}
	\EEA
	where $\sign(\cdot)$ and $\max\{\cdot,\cdot\}$ functions are applied element-wise to the vector arguments. To compute $\wg^{t+1}$ from $\{\ww_k^{t+1} \,:\, k\in \S^t \}$ the aggregator starts with $\wg = \ww_{mean} := \mathtt{Mean}\{ \ww_k^t\, : \,k \in \S^t\}$ and runs the following two step iterative procedure until $\wg$ converges:
	\BEA
	\begin{array}{l}
	\vv_k \gets \mathtt{Soft\_Thresholding}(\ww_k^{t+1} - \wg, \,\delta ), \,\forall \,k \in \S^t, \\
	\wg  \gets  \ww_{mean} -\mathtt{Mean}\{ \vv_k : k \in \S^t \}.
	\end{array}\nonumber
	\EEA
	
{\bf Hybridization via the Unified Fed+ Framework with Layer-specific $\Psi$}: The unification of aggregation methods through a single formulation allows for seamlessly combining different methods of aggregation and personalization to different layers in training deep neural networks. For example, initial layers may use FedAvg+, while final layers may benefit from FedCoMed+. Also, the level of personalization can be controlled by setting layer-specific $\delta$.

{\bf Deriving Existing Algorithms from Fed+:} ~Many federated learning methods fit into the Fed+ framework and can be obtained by setting the parameters in Algorithm~\ref{alg:FP} appropriately, as summarized in Table \ref{tab:tab}.

{
\newcommand{\mrtt}[1]{\betaltirow{2}{*}{#1}} 
\newcommand{\mrt}[1]{\betaltirow{3}{*}{#1}} 
\newcommand{\mcf}[1]{\betalticolumn{4}{c}{#1}}
\newcommand{\mcs}[1]{\betalticolumn{6}{c}{#1}}
\newcommand{\mct}[1]{\betalticolumn{3}{c}{#1}}
\newcommand{\note}{${}^\dagger$}
\begin{table*}
    \centering
    \begin{tabular}{l||l|c|c|l}
    
    Method & Aggregation Function ${\cal A}$ & $\lambda$ & $\sigma$ & other remarks\\
    \hline \hline
     Local SGD without  FL   & ~NA &  $ \lambda = 0$ & $\sigma = 0$ & NA 
     \\
     FedAvg    & ~$\mathtt{Mean}$ & $\lambda =1$& $\sigma = 0$& $E_k=E,\forall  k$
     \\
     RFA & ~$ \mathtt{Geometric\,\,Median}$ & $\lambda =1$ & $\sigma = 0$ & $E_k=E,\forall  k$
     \\
     Coordinatewise median & ~$\mathtt{Coordinatewise\,\,Median}$ & $\lambda =1$& $\sigma = 0$& $E_k=E,\forall  k$ 
     \\
    FedProx & ~$\mathtt{Mean}$ & $\lambda =1$ & $\sigma >0$ & $\vv_k^t  = \zer, \forall k$ 
     \\
    FedAvg+ & ~$\mathtt{Mean}$ & $\lambda =0$ & $\sigma >0$ & $\vv_k^t  \neq \zer, \forall k$ 
     \\     
    FedGeoMed+ & ~$\delta-\mathtt{Geometric\,\,Median}$ & $\lambda =0$ & $\sigma >0$ & $ \delta > 0$ 
     \\     
    FedCoMed+ & ~$\delta-\mathtt{Coordinatewise\,\,Median}$ & $\lambda =0$ & $\sigma >0$ & $\delta > 0$ 
     \\

\hline\hline
    \end{tabular}
    \caption{Deriving Existing and Proposed Algorithms from Fed+. (In FedProx, $\vv_k^t  = \zer, \forall k$  corresponds to choosing $\Psi$ to be $0$ at $\zer$ and $+\infty$ elsewhere). The notation $\approx^{\delta}$ refers to a $\delta$-approximation.  }
    \label{tab:tab}
\end{table*}
}

%%%%====================================================================================

%%%%%%%%%%%%%%%%%%%%%%%%%%%%%%%%%%%%%%%%%%%%
\subsection{Convergence Analysis of Fed+}
%%%%%%%%%%%%%%%%%%%%%%%%%%%%%%%%%%%%%%%%%%%%

The convergence properties and fixed points of the Fed+ algorithm are presented next. The parameters $\sigma>0$, $\delta >0$, and $\eta >0$ are tunable unless specified otherwise. For the rest of this section, we will use the following setting for the parameters in Algorithm~\ref{alg:FP}: (i) $\Psi: \RR^d \to [0,\infty]$ is any convex function with an easy to compute proximal operator and  (ii) the personalization vector $\vv_k$ is set as in eqn. \eqref{eq:zk_rp_def}. 
To implement the aggregation step $\wg \gets \A(\ww_1,\ldots,\ww_\K)$ for a general choice of $\Psi$, we propose the following iterative procedure initialized with $\wg = \ww_{mean} := \mathtt{Mean}\{ \ww_1,\ldots, \ww_\K\}$:
\BEA\begin{array}{l}
	\vv_k ~ \gets  ~ \prox^{\frac{1}{\sigma}}_{\Psi}(\ww_k  - \wg),~~ \,k = 1,\ldots,\K,  \\
	\wg ~ \gets ~ \ww_{mean} -\mathtt{Mean}\{ \vv_1 ,\ldots, \vv_K \},  
	\end{array}
	%\wg ~& \gets &~ \ww_{mean} \,- \,\mathtt{Mean}\big\{\, \prox^{\rho}_{\Psi}(\ww_1  - \wg) , \, \cdots, \, \prox^{\rho}_{\Psi}(\ww_\K  - \wg) \,\big\}. 
	\label{eq:xhat_iter_phi}	
\EEA
The above setting  gives rise to the following useful property:
\BEA
	(\vv_1^{t}, \ldots, \vv_{\K}^t,\wg^{t}) = \argmin_{ \V, \,\wg} H_{\sigma}(\W^t,\V,\wg),~t\ge1. \label{eq:Z_w_update}
\EEA
To analyze Fed+, we make the following smoothness assumption:
\begin{ass}\label{ass:fk}
For each $k = 1,2,\ldots,\K$, $f_k:\RR^d \to \RR$ is differentiable and the gradient $\nabla f_k$ is Lipschitz continuous with constant $L_f$, i.e.,
\[ 
    \| \nabla f_k(\ww) - \nabla f_k(\ww') \|_2 \le L_f \| \ww - \ww' \|_2, ~\forall \ww , \ww' \in \RR^d.
\] 
\end{ass}	
\begin{prop}\label{prop:grad_decrease}
Under Assumption~\ref{ass:fk} and the stepsize choice $\eta = 1/L_f$, the following holds for Fed+:	$\forall k \in \S_t,$
\BEA
F_k( \ww_k^{t+1}; \vv_k^{t},\wg^{t}) ~\le ~  F_k( \ww_k^{t}; \vv_k^{t},\wg^{t})  - \frac{\| \nabla F_k(\ww_k^{t}; \vv_k^{t},\wg^{t}) \|_2^2}{2(L_f+\sigma)} , \label{eq:Fk_decrease}
\EEA
where $F_k$ is defined in \eqref{eq:Fk_def} and the gradient is w.r.t. $\ww_k$.
\end{prop}
We define the federated training objective for our set-up as:
\BEA
F_\sigma(\W) ~&:=&~ \argmin_{ \V, \,\wg} H_{\sigma}(\W,\V,\wg)
%\\&=& \frac{1}{\K}\sum_{k=1}^{\K}  \left[\, f_k(\ww_k) \,+   \hat{\Psi}_{\frac{1}{\sigma}}(\ww_k- \A(\W)) \, \right].
\label{eq:Fmu}
\EEA
Now, combining the relation \eqref {eq:Z_w_update} with \eqref{eq:Fk_decrease} we derive the following convergence result for Fed+:
\begin{thm}\label{thm:thm_new_prob_det}
	Assume that $H_{\sigma}$ in \eqref{eq:FPz} is bounded from below, parties are sampled with equal probability. Then, under Assumption~\ref{ass:fk} and the stepsize choice $\eta = 1/L_f$, the following holds for Fed+:
	\BEA
	\lim_{t\to \infty} \prb \left[ \sum_{k=1}^\K \| \nabla F_k(\ww_k^{t};\vv_k^{t},\wg^{t}) \|^2_2 \right] = 0, \label{eq:exp_grad_zero}
	\EEA
	where the expectation is with respect to the random subsets $\S^t$, ~$t\ge 0$. \\
	Moreover, the federated objective $F_\sigma(\W^t)$ monotonically decreases with round $t$ and converges to a value $\hat{F}_\sigma \ge \min_{\W}F_\sigma(\W).$ 
	Additionally, if the $f_k$'s are convex, all parties are active in every round, and the level set $\{ (\W,\V,\wg) \,:\,H_{\sigma}(\W,\V,\wg) \le  H_{\sigma}(\W^0,\V^0,\wg^0)\}$ is compact, then $ \lim_{t\to \infty}F_\sigma(\W^t) = \min_{\W}F_\sigma(\W)$ and the rate of convergence is $\mathcal{O}(1/t)$.
\end{thm}

\subsection{Fixed Points of Fed+}
Here, we present the characterization of the fixed points of Fed+ algorithm to gain insight on the kind of personalized solution it offers. Before proceeding further we make the following assumption:
\begin{ass}\label{ass:exact_prox_convex}
	For each $k =1,\ldots, \K$, $f_k$ is convex, all the parties actively participate in every round of the federating learning process, and the Local-Solve subroutine in Fed+ returns $\ww_k^{t+1}$ as the exact minimizer of $F_k(\cdot\,;\vv_k^{t},\wg^{t})$. %, i.e., 
%	\BEA \ww_k^{t+1} ~=~  \prox_{f_k}^{\frac{1}{\sigma}}(\vv_k^{t}+\wg^t) ~:=~ \argmin_{\ww_k} f_k(\ww_k) + \frac{\sigma}{2}\|\ww_k - (\vv_k^{t}+\wg^t)\|_2^2, ~\forall \,t \ge 0, \forall k . \label{eq:prox_fk}
%	\EEA
\end{ass}
We define $\hat{f}_k:\RR^d \to \RR$ to be the Moreau envelope of $f_k$ with smoothing parameter $(1/\sigma)$, i.e., 
\BEA
\hat{f}_k(\vv) ~:=~ \min_{\ww_k \in \RR^d } ~f_k(\ww_k) + \frac{\sigma}{2} \|\ww_k-\vv\|_2^2, ~~\forall \vv \in \RR^d.
\EEA
The fixed-point characterization of Fed+ under Assumption~\ref{ass:exact_prox_convex} is thus:
\begin{thm}\label{thm:FixedPointThm} 
	Consider the Fed+ algorithm for solving problem \eqref{eq:FPz} under Assumption~\ref{ass:exact_prox_convex}. Let $(\W^*,\V^*,\wg^*)$ be a fixed point of Fed+ and $\zz_k^* := \wg^* + \prox^{\frac{1}{\sigma}}_{\Psi} ( \ww_k^* - \wg^*)$. Then, the following conditions are satisfied:
	\BEA 
	\frac{1}{\K}\sum_{k=1}^\K \nabla \hat{f}_k(\zz_k^*)  = 0, ~~~
	\ww_k^*  = \zz_k^* - \frac{1}{\sigma} \nabla \hat{f}_k(\zz_k^*) , \forall k. \label{eq:FixPointEq}
	\EEA
%	Moreover, the set of fixed points is precisely the set of optimal solution of \eqref{eq:F_H}. 
\end{thm}	
Now, with the help of the above Theorem, we analyze two extreme choices for $\Psi$ in part (a) \& (b) of the following Corollary: 
\begin{cor}\label{cor:fixpoint_coro}
	Consider the Fed+ algorithm under Assumption~\ref{ass:exact_prox_convex}. Let $(\W^*,\V^*,\wg^*)$ be a fixed point of Fed+. Then, the following are true: \\
	(a) If we choose  ~$\Psi \equiv 0$ (i.e. $\vv_k^t = \ww_k^t-\wg^t$) in Fed+, then 
	\BEA
	\ww_k^* ~\in~ \argmin_{\ww}f_k(\ww),~k=1,\ldots,\K. \label{eq:fixa}
	\EEA
	(b) If Fed+ sets ~$\Psi(\ww) = 0$ iff $\ww = \zer$ and $+\infty$ otherwise (i.e. $\vv_k^t = \zer$), then 
		\BEA
			\frac{1}{\K}\sum_{k=1}^\K \nabla \hat{f}_k(\wg^*)  = 0,~~~ \ww_k^* = \wg^* - \frac{1}{\sigma} \nabla \hat{f}_k(\wg^*),~ \forall k. \label{eq:fixb}
		\EEA
	(c) If Fed+ employs ~$\Psi(\ww) = \frac{\sigma \delta}{2}\|\ww\|_2^2 $ leading to $\vv_k^t = [1+\delta]^{-1} ( \ww_k^t -\wg^t)$, then 
		\BEA
			\ww_k^* = \wg^* - \left(\frac{1+\delta}{\sigma\delta}\right) \nabla \hat{f}_k\left(\frac{\ww_k^* + \delta \wg^*}{1+\delta}\right), ~k=1,\ldots,\K, %\nonumber \\ \hspace{2cm} \mbox{where}~~ \wg^* = \frac{1}{\K}\sum_{k=1}^\K \ww_k^*. \nonumber
		\EEA
		where $\wg^* = \frac{1}{\K}\sum_{k=1}^\K \ww_k^*$. 
\end{cor}

%%%%====================================================================================

%%%%====================================================================================
\section{Experiments}
%%%%====================================================================================

\subsection{Performance Comparison on Personalized Datasets}
To test  the robustness as well as the personalization quality of Fed+ and the baseline methods, we  create two  variants of the MNIST dataset: MNIST-robust and MNIST-personal, each with two different federation sizes, $N=10$ and $N=50$, as well as a personalized synthetic regression problem. To create the robust and personal variants, we first partitioned the MNIST dataset equally into $\K$ parties. Then we transform the input distribution for 10\% (20\% for $\K=50$) of the parties by taking negative of the images. Further, for every party we choose 2 different class labels and add Laplacian noise to the images corresponding to those classes to create the personalized dataset. In the synthetic regression dataset ($\K =10$), a sample $(\mathbf{x},y)$ for a party $k$ is generated through model: $ y = \ww_k^{T}\mathbf{x} + v$, where $\ww_k \in \RR^{1000}$, $v \sim \N(0,2)$ and $\mathbf{x} \sim \N(\mu_k,\Sigma)$ with diagonal covariance matrix given by $\Sigma_{j,j}= \mbox{mod}(j,50)^{-1.1}$; $\mu_k \sim \N(0,0.5)$. Party specific weight vectors $\{ \ww_k \in \RR^{1000} \}_{k=1}^{\K-1}$ are generated by adding Laplacian noise (with scale =0.5) to a fixed $\bar{\ww} \sim \N(0,5\eye_{1000})$. We generate $\ww_\K$ similarly but with a different $\bar{\ww}\sim \N(0,50\eye_{1000})$ to make the setting robust.% the robustness aspect of the FL methods. 

We train logistic regression classifiers on the robust and personalized variants of the MNIST dataset and a linear regression model on the Synthetic one. Data is randomly split for each local party into an 50\% training set and a 50\% test set. We report average performance on the test set after running each method 5 times with different random seeds.
Each party's dataset consists of 100 samples in the Synthetic-regression case. The number of selected parties per round is $K=10$; the batch size is 20 for MNIST datasets and 10 for the synthetic-regression dataset. We set the learning rate $\eta$ to 0.0001 for synthetic and 0.02 for MNIST variants. The regularization constant $\sigma$ is chosen to be 15 ($\sigma=1$ for synthetic) for the robust and personalized MNIST datasets. For all experiments, we fix the number of local iterations per round by setting $E_k = 20$ and report performance after $T=500$ rounds of training. %For the MNIST datasets, we do not consider the parties whose images were transformed by taking negative

We  compare the performance of Fed+ with  FL methods such as the non-personalized Scaffold \citep{2019SCAFFOLD}, as well as the personalized FL algorithms  pFedMe \citep{moreau},
perFedAvg \citep{perFedAvg} and APFL \citep{deng2020adaptive}. The results are compiled in Table \ref{tab:performance_2}. Note that the MNIST problems measure accuracy and as such higher is better, while the regression problem measures error and thus lower is better. 

We observe that the robust FedGeoMed+ outperforms both the non-personalized and mean-aggregation personalized methods by a significant margin. Our FedCoMed+, while performing poorly on the personalized MNIST-robust datasets, is a close second place on the synthetic regression problem.

\begin{table*}[t]
	\centering	
	\small
	\caption{Average performance across recent FL methods on personalized FL datasets. For MNIST, higher (accuracy) is better; for synthetic-regression, lower (error) is better.
	We provide results for the non-personalized  Scaffold and three personalized methods   (left) along with  our Fed+ algorithms: FedAvg+, FedGeoMed+ and FedCoMed+ (right). }\label{tab:performance_2}
	 \setlength{\tabcolsep}{4pt}
	\begin{tabular}{l|cccc|ccc}
		Dataset \!\!&\! Scaffold \!&\! pFedMe \!&\! perFedAvg \!&\! APFL \!&\! FedAvg+ \!&\! FedGeoMed+ \!&\! FedCoMed+ \\ \hline \hline
 %   MNIST & 81.0 & 81.0 & 80.9 & 82.4 & 82.1 & 80.5 & 82.0 & 82.1 & 82.0 \\ \hline
 MNIST-robust-N10 & 81.4 & 89.4 & 87.3 & 88.3 & 87.0 & {\bf 91.5} & 80.5  \\ \hline
 MNIST-robust-N50 & 82.3 & 85.5 & 85.7 & 81.3 & 86.7 & {\bf 91.4} & 79.4 \\ \hline
 MNIST-personal-N10 & 72.0 & 72.3 & 70.6 & 75.7 & 71.9 & {\bf 78.3} & 66.7 \\ \hline
 MNIST-personal-N50 & 63.8 & 66.3 & 68.9 & 72.7 & 69.4 & {\bf 76.2 } & 52.3 \\ \hline
 %   FEMNIST-robust-N10 & & & & & & &  \\ \hline
%        FEMNIST-robust-N50 & & & & & & &  \\ \hline
    Synthetic-regression  & 2764 & 4268 & 2780 & 1606 & 1966 & {\bf 1048} & 1074\\ \hline
	\end{tabular}
\end{table*}

\begin{figure*}[hbt!]
	\centering
	\includegraphics[width=1\textwidth]{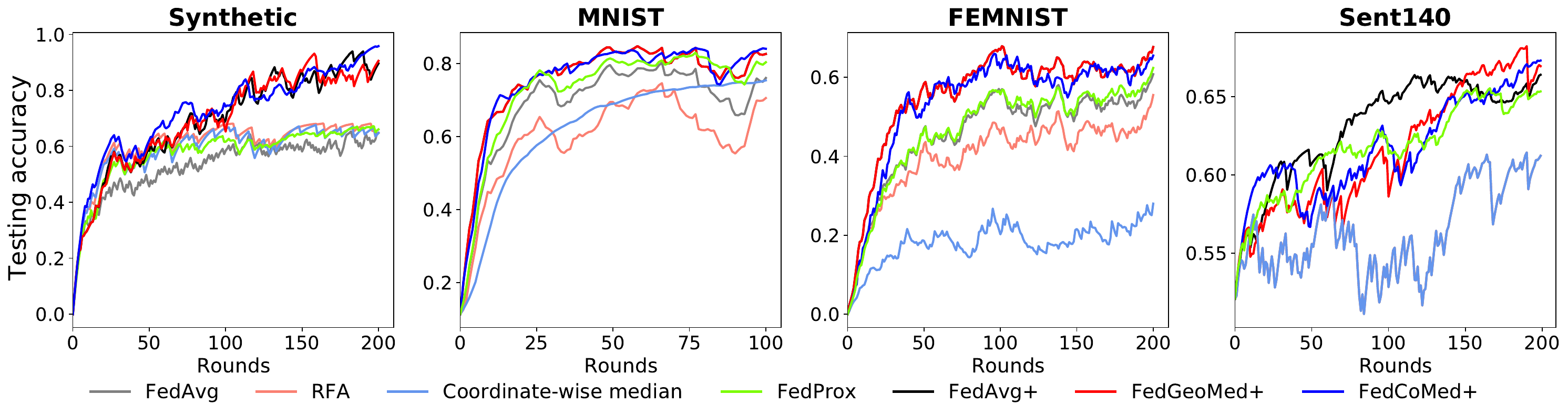}
	\caption{Performances of  FedAvg+, FedGeoMed+ and FedCoMed+ are superior to those of the baselines.}\label{fig:benchmarks}
\end{figure*}

\subsection{Results on Standard Federated Learning Datasets}
We also test our methods and the main non-personalized methods on a  set of synthetic and non-synthetic datasets from \citet{li2020federated} and the LEAF set of \citet{caldas2018leaf}. Since our FedAvg+ is comparable to several of the recent personalized FL methods, it serves as an indicator of how mean aggregation-based personalized models fare on these standard datasets.

To generate non-identical synthetic data, we follow a similar setup to that of \citet{li2020federated}, additionally imposing heterogeneity among parties. In particular, for each party $k$, we generate samples $(X_k, Y_k)$ according to the model $y=\arg\max(\textrm{softmax}(Wx+b))$, $x \in \RR^{60}, W \in \RR^{10 \times 60}, b \in \RR^{10}$.  We model $W_k \sim \N(u_k, 1)$, 
$b_k \sim \N(u_k, 1)$, $u_k \sim \N(0, \zeta)$; $x_k \sim \N(v_k, \Sigma)$, where the covariance matrix $\Sigma$ is diagonal with $\Sigma_{j,j}=j^{-1.2}$. Each element in the mean vector $v_k$ is drawn from $\N(B_k, 1), B_k \sim \N(0, \beta)$. Therefore, $\zeta$ controls how much the local models differ from each other and $\beta$ controls how much the local data at each party differs from that of other parties. In order to better characterize statistical heterogeneity and study its effect on convergence, we choose $\zeta=1000$ and $\beta=10$.   There are $\K=30$ parties in total and the number of samples on each party follows a power law. 

The hyperparameters are the same as those of  \citet{li2020federated} and use their reported best $\mu$ for their algorithm FedProx. 
 MNIST~\citep{lecun1998gradient} with multinomial logistic regression. To impose statistical heterogeneity, we distribute the  data  among $\K=$1,000 parties such that each party has samples of only one digit and the number of samples per party follows a power law. The input  is a flattened 784-dimensional (28 $\times$ 28) image, and the output is a class label between 0 and 9. 
We also include the 62-class Federated Extended MNIST~\citep{cohen2017emnist,caldas2018leaf} (FEMNIST) of \citet{li2020federated}. Heterogeneous data partitions are generated by subsampling $10$ lower case characters (`a'-`j') from EMNIST  and distributing only 5 classes to each party, with $\K=$200 parties in total. The input is a flattened 784-dimensional (28 $\times$ 28) image, and the output is a class label between 0 and 9. 
To address non-convex settings, we consider  sentiment analysis  on tweets from Sentiment140~\citep{go2009twitter} (Sent140) with a two layer LSTM binary classifier containing 256 hidden units with pretrained 300D GloVe embedding~\citep{pennington2014glove}.
Each twitter account corresponds to a party with $\K=$772  in total. The model takes as input a sequence of 25 characters, embeds each  into a 300-dimensional space using Glove and outputs one character per training sample after 2 LSTM layers and a densely-connected layer. We consider the highly heterogeneous setting where there are $90\%$ stragglers; see \citet{li2020federated} for  details. %Details of the synthetic dataset and implementation can be found in the supplementary material.

Data is randomly split for each local party into an 80\% training set and a 20\% testing set. The number of selected parties per round is 10 and the batch size is 10 for all experiments on all datasets. The neural network models for all datasets are the same as those of \citet{li2020federated}. Learning rates are 0.01, 0.03, 0.003 and 0.3 for synthetic, MNIST and FEMNIST and Sent140 datasets, respectively. The experiments used a fixed regularization parameter $\sigma=0.01$ for each party's Local-Solve and the parameter $\delta$ is set to $0.001, 0.1$ and $0.1$ for FedAvg+, FedGeoMed+ and FedCoMed+ methods, respectively. On the Sent140 dataset, we found that initializing the local model to a mixture model (i.e. setting $\lambda = 0.001$ instead of the default $\lambda = 0$) at the beginning of every Local-Solve subroutine for each party gives the best performance. %The mixture model is computed using a convex combination of each party's local model and latest global model with weight $0.001$. 
We simulate the federated learning setup (1 aggregator $\K$ parties) on a commodity-hardware machine with 16 Intel$^\text{\textregistered}$ Xeon$^\text{\textregistered}$ E5-2690 v4 CPU and 2 NVIDIA$^\text{\textregistered}$ Tesla P100 PCIe GPU.

In Figure \ref{fig:benchmarks}, we illustrate the test performance of the  baseline algorithms FedAvg, FedProx, RFA, coordinate-wise median  and Fed+. FedAvg+ is comparable and to and thus represents the performance of the recent personalized methods.  The baseline robust   algorithms  perform the worst on these non-IID data sets.   Fed+ often  speeds up the learning convergence, as shown in Figure \ref{fig:benchmarks} and improves  performance on these  datasets by $28.72\%$, $6.24\%$, $11.32\%$ and $13.89\%$, resp. In particular, the best Fed+ algorithm can improve the most competitive implementation of the baseline  FedProx on these four datasets by $9.90\%$ on average. FedAvg+, and hence many standard personalized FL methods,   achieves similar performance to FedGeoMed+ on  MNIST and FEMNIST, but but fails to outperform the robust variants of Fed+,  FedGeoMed+ and FedCoMed+, on the synthetic and Sent140 datasets, highlighting the benefit of  robust statistics.

We  also evaluate the impact of increasing the number of parties in training on  test accuracy. On the synthetic dataset, average test accuracy improves from $70.22\%$ to $90.73\%$ to $98.03\%$ when the number of parties participating in training goes from $\K=3$ to $\K=15$ to $\K=30$. The average is taken over the three Fed+ variants: FedAvg+, FedGeoMed+ and FedCoMed+. On  MNIST, average accuracies of Fed+  are $69.80\%$, $81.34\%$, and $83.36\%$ when the number of parties in training goes from $\K=100$ to $\K=500$ to $\K=1000$, resp.
On  FEMNIST, average accuracies  of Fed+  are $25.16\%$, $68.71\%$, and $78.66\%$ when the number of parties in training goes from $\K=20$ to $\K=100$ to $\K=200$, resp.
On the Sent140 dataset, average accuracies of Fed+  are $57.13\%$, $60.77\%$, and $65.43\%$ when the number of parties goes from $\K=77$ to $\K=386$ to $\K=772$, resp.
This shows that the  benefit of using Fed+ increases  as the number of parties increases.

%%%%====================================================================================
\section{Conclusion}
%%%%====================================================================================

Fed+~has been designed to better handle the heterogeneity inherent in federated settings: the lack of IID data, the need for robustness to outliers and stragglers, and the requirement to perform well on party-specific data. 
The Fed+ class of methods  unifies numerous algorithms through a formulation that allows for robust ways of aggregating the local models whilst keeping the structure of  local computation intact. 
We provide convergence guarantees for  Fed+  for  convex and non-convex loss functions, robust  aggregation, and for the case of stragglers.
Probably the most promising extension of this work would be an in-depth exploration of   neural network layer-specific aggregation functions as made possible through the Fed+ formulation.

%%%%=========================================================================
\bibliographystyle{named}
\bibliography{references_FedPlus}

\begin{thebibliography}{}

\bibitem[\protect\citeauthoryear{Beck}{2015}]{Beck_alternate_minimization_convergence}
Amir Beck.
\newblock On the convergence of alternating minimization for convex programming
  with applications to iteratively reweighted least squares and decomposition
  schemes.
\newblock {\em SIAM Journal on Optimization}, 25(1):185--209, 2015.

\bibitem[\protect\citeauthoryear{Caldas \bgroup \em et al.\egroup
  }{2018}]{caldas2018leaf}
Sebastian Caldas, Peter Wu, Tian Li, Jakub Kone{\v{c}}n{\`y}, H~Brendan
  McMahan, Virginia Smith, and Ameet Talwalkar.
\newblock {LEAF}: A benchmark for federated settings.
\newblock {\em arXiv preprint arXiv:1812.01097}, 2018.

\bibitem[\protect\citeauthoryear{Charles and
  Konecn{\'y}}{2020}]{Charles2020Local}
Zachary Charles and Jakub Konecn{\'y}.
\newblock On the outsized importance of learning rates in local update methods.
\newblock {\em ArXiv}, abs/2007.00878, 2020.

\bibitem[\protect\citeauthoryear{Cohen \bgroup \em et al.\egroup
  }{2017}]{cohen2017emnist}
Gregory Cohen, Saeed Afshar, Jonathan Tapson, and Andre Van~Schaik.
\newblock {EMNIST}: Extending {MNIST} to handwritten letters.
\newblock In {\em 2017 International Joint Conference on Neural Networks
  (IJCNN)}, pages 2921--2926. IEEE, 2017.

\bibitem[\protect\citeauthoryear{Deng \bgroup \em et al.\egroup
  }{2020}]{deng2020adaptive}
Yuyang Deng, Mohammad~Mahdi Kamani, and Mehrdad Mahdavi.
\newblock Adaptive personalized federated learning.
\newblock {\em arXiv preprint arXiv:2003.13461}, 2020.

\bibitem[\protect\citeauthoryear{Fallah \bgroup \em et al.\egroup
  }{2020}]{perFedAvg}
Alireza Fallah, Aryan Mokhtari, and A.~Ozdaglar.
\newblock Personalized federated learning with theoretical guarantees: A
  model-agnostic meta-learning approach.
\newblock In {\em NeurIPS}, 2020.

\bibitem[\protect\citeauthoryear{Go \bgroup \em et al.\egroup
  }{2009}]{go2009twitter}
Alec Go, Richa Bhayani, and Lei Huang.
\newblock Twitter sentiment classification using distant supervision.
\newblock {\em CS224N project report, Stanford}, 1(12):2009, 2009.

\bibitem[\protect\citeauthoryear{Hanzely and
  Richt{\'a}rik}{2020}]{KAUSTLocalGlobal}
Filip Hanzely and Peter Richt{\'a}rik.
\newblock Federated learning of a mixture of global and local models.
\newblock {\em ArXiv}, abs/2002.05516, 2020.

\bibitem[\protect\citeauthoryear{Hanzely \bgroup \em et al.\egroup
  }{2021}]{unified_personalized}
Filip Hanzely, Boxin Zhao, and Mladen Kolar.
\newblock Personalized federated learning: A unified framework and universal
  optimization techniques, 2021.

\bibitem[\protect\citeauthoryear{Karimireddy \bgroup \em et al.\egroup
  }{2019}]{2019SCAFFOLD}
Sai~Praneeth Karimireddy, S.~Kale, M.~Mohri, S.~Reddi, S.~Stich, and A.~T.
  Suresh.
\newblock {SCAFFOLD}: {S}tochastic controlled averaging for on-device federated
  learning.
\newblock {\em ArXiv}, abs/1910.06378, 2019.

\bibitem[\protect\citeauthoryear{Kone{\v{c}}n{\`y} \bgroup \em et al.\egroup
  }{2015}]{konevcny2015federated}
Jakub Kone{\v{c}}n{\`y}, Brendan McMahan, and Daniel Ramage.
\newblock Federated optimization: {D}istributed optimization beyond the
  datacenter.
\newblock {\em arXiv preprint arXiv:1511.03575}, 2015.

\bibitem[\protect\citeauthoryear{Kone{\v{c}}n{\`y} \bgroup \em et al.\egroup
  }{2016}]{konevcny2016federated}
Jakub Kone{\v{c}}n{\`y}, H~Brendan McMahan, Felix~X Yu, Peter Richt{\'a}rik,
  Ananda~Theertha Suresh, and Dave Bacon.
\newblock Federated learning: {S}trategies for improving communication
  efficiency.
\newblock {\em arXiv preprint arXiv:1610.05492}, 2016.

\bibitem[\protect\citeauthoryear{LeCun \bgroup \em et al.\egroup
  }{1998}]{lecun1998gradient}
Yann LeCun, L{\'e}on Bottou, Yoshua Bengio, and Patrick Haffner.
\newblock Gradient-based learning applied to document recognition.
\newblock {\em Proceedings of the IEEE}, 86(11):2278--2324, 1998.

\bibitem[\protect\citeauthoryear{Li \bgroup \em et al.\egroup
  }{2020a}]{li2020federated}
Tian Li, Anit~Kumar {Sahu}, Manzil {Zaheer}, Maziar {Sanjabi}, Ameet
  {Talwalkar}, and Virginia {Smith}.
\newblock Federated optimization in heterogeneous networks.
\newblock {\em Proceedings of Machine Learning and Systems}, 2:429--450, 2020.

\bibitem[\protect\citeauthoryear{Li \bgroup \em et al.\egroup
  }{2020b}]{ConvFedAvg}
Xiang Li, Kaixuan Huang, Wenhao Yang, Shusen Wang, and Zhihua Zhang.
\newblock On the convergence of {F}ed{A}vg on non-iid data.
\newblock In {\em International Conference on Learning Representations}, volume
  Arxiv, abs/1907.02189, 2020.

\bibitem[\protect\citeauthoryear{Li \bgroup \em et al.\egroup }{2021}]{ditto}
Tian Li, Shengyuan Hu, Ahmad Beirami, and Virginia Smith.
\newblock Ditto: Fair and robust federated learning through personalization.
\newblock In Marina Meila and Tong Zhang, editors, {\em Proceedings of the 38th
  International Conference on Machine Learning, {ICML}}, volume 139 of {\em
  Proceedings of Machine Learning Research}, pages 6357--6368. {PMLR}, 2021.

\bibitem[\protect\citeauthoryear{Malinovsky \bgroup \em et al.\egroup
  }{2020}]{Malinovsky2020}
G.~Malinovsky, D.~Kovalev, E.~Gasanov, Laurent Condat, and Peter Richt{\'a}rik.
\newblock From local {SGD} to local fixed point methods for federated learning.
\newblock {\em ICML}, Arxiv, abs/2004.01442, 2020.

\bibitem[\protect\citeauthoryear{{Mansour} \bgroup \em et al.\egroup
  }{2020}]{3personalized}
Yishay {Mansour}, Mehryar {Mohri}, Jae {Ro}, and Ananda {Theertha Suresh}.
\newblock {Three Approaches for Personalization with Applications to Federated
  Learning}.
\newblock {\em arXiv e-prints}, page arXiv:2002.10619, February 2020.

\bibitem[\protect\citeauthoryear{McMahan \bgroup \em et al.\egroup
  }{2017}]{mcmahan2017communication}
Brendan McMahan, Eider Moore, Daniel Ramage, Seth Hampson, and Blaise~Aguera
  y~Arcas.
\newblock Communication-efficient learning of deep networks from decentralized
  data.
\newblock In {\em Artificial Intelligence and Statistics}, pages 1273--1282.
  PMLR, 2017.

\bibitem[\protect\citeauthoryear{Pathak and
  Wainwright}{2020}]{Pathak2020FedSplit}
Reese Pathak and M.~Wainwright.
\newblock {F}ed{S}plit: {A}n algorithmic framework for fast federated
  optimization.
\newblock {\em ArXiv}, abs/2005.05238, 2020.

\bibitem[\protect\citeauthoryear{Pennington \bgroup \em et al.\egroup
  }{2014}]{pennington2014glove}
Jeffrey Pennington, Richard Socher, and Christopher~D Manning.
\newblock Glove: {G}lobal vectors for word representation.
\newblock In {\em Proceedings of the 2014 conference on empirical methods in
  natural language processing (EMNLP)}, pages 1532--1543, 2014.

\bibitem[\protect\citeauthoryear{Pillutla \bgroup \em et al.\egroup
  }{2019}]{pillutla2019robust}
Krishna Pillutla, Sham~M Kakade, and Zaid Harchaoui.
\newblock Robust aggregation for federated learning.
\newblock {\em arXiv preprint arXiv:1912.13445}, 2019.

\bibitem[\protect\citeauthoryear{T.~Dinh \bgroup \em et al.\egroup
  }{2020}]{moreau}
Canh T.~Dinh, Nguyen Tran, and Josh Nguyen.
\newblock Personalized federated learning with moreau envelopes.
\newblock In H.~Larochelle, M.~Ranzato, R.~Hadsell, M.~F. Balcan, and H.~Lin,
  editors, {\em Advances in Neural Information Processing Systems}, volume~33,
  pages 21394--21405. Curran Associates, Inc., 2020.

\bibitem[\protect\citeauthoryear{Yin \bgroup \em et al.\egroup
  }{2018}]{pmlr-v80-yin18a}
Dong Yin, Yudong Chen, Ramchandran Kannan, and Peter Bartlett.
\newblock {B}yzantine-robust distributed learning: {T}owards optimal
  statistical rates.
\newblock In Jennifer Dy and Andreas Krause, editors, {\em Proceedings of the
  35th International Conference on Machine Learning}, volume~80 of {\em
  Proceedings of Machine Learning Research}, pages 5650--5659. PMLR, 10--15 Jul
  2018.

\bibitem[\protect\citeauthoryear{Zhang \bgroup \em et al.\egroup
  }{2021}]{localcommonsoln}
Michael Zhang, Karan Sapra, Sanja Fidler, Serena Yeung, and Jose~M. Alvarez.
\newblock Personalized federated learning with first order model optimization.
\newblock In {\em International Conference on Learning Representations}, 2021.

\end{thebibliography}
%%%%=========================================================================

%%%%%%%%%%%%%%%%%%%%%%%%%%%%%%%%%%%%%%%%%%%%%%%%%%%%%%%%%%%%%%%%%%%%%%%%%%%%%%%%%%%
%%%%%%%%%%%%%%%%%%%%%%%%%%%%%%%%%%%%%%%%%%%%%%%%%%%%%%%%%%%%%%%%%%%%%%%%%%%%%%%%%%%%
\newpage
%\nolinenumbers
\appendix
\begin{centering}
	{\LARGE{\bf Appendix}} \\ \vspace{1cm}
\end{centering}
%\linenumbers
%\resetlinenumber
Here, we prove all the propositions and the theorems stated in the paper. %We also provide additional details on the experimental set-up.

\subsection{Proof of Proposition~\ref{prop:F_grad_step}}
The gradient descent iteration for the function $F_k(\cdot\,;\vv_k^{t},\wg^{t})$ with stepsize $\eta' := \frac{\eta}{1+\sigma\eta}$ is given by
\BEA
\ww_k &\gets&  \ww_k - \eta'[ \,\nabla f_k (\ww_k) + \sigma( \ww_k - \vv_k^{t}-\wg^t) \,] \nonumber \\
& =& (1 - \sigma \eta') \ww_k^t - \eta' \nabla f_k(\ww_k) +(\sigma \eta') \left[\vv_k^{t} + \wg^t\right] \nonumber \\
& =&  \left(\frac{1}{1+\sigma\eta}\right) \left[ \, \ww_k - \eta \nabla f_k(\ww_k)\,\right] \,+\, \left(\frac{\sigma\eta}{1+\sigma\eta}\right) \left[\vv_k^{t} + \wg^t\right]. \nonumber
\EEA
Thus, we have the local update of the form \eqref{eq:alg_local_update} in Fed+ algorithm where $\kappa := \frac{1}{1+\sigma\eta}$.

%%%%----------------------------------------------------------

\subsection{Proof of Proposition~\ref{prop:grad_decrease}}
Let us first recall the following well-know descent lemma \cite{Beck_alternate_minimization_convergence} for functions with Lipschitz continuous gradient. 
\begin{lem}\label{lem:descent_lemma}
	Let $f: \RR^d \to \RR$ be continuously differentiable and $\nabla f$ be Lipschitz continuous with constant $L >0$. Then, the following holds:
	\BEA f\left( \,\ww - \frac{1}{L} \nabla f(\ww)\,\right) \le f(\ww) - \frac{1}{2L} \| \nabla f(\ww)\|_2^2,~\forall \,\ww \in \RR^d. \nonumber
	\EEA
	%	If $\gamma \in \left(0,\frac{2}{L}\right)$ then $f\left( \,\ww - \gamma \nabla f(\ww)\,\right) ~\le~ f(\ww)$.
\end{lem}
From Proposition~\ref{prop:F_grad_step}, we know that the local update \eqref{eq:alg_local_update} in Fed+ algorithm is a gradient descent iteration with learning rate $\eta' = \frac{\eta}{1+\sigma\eta} = \frac{1}{L_f+\sigma}$ applied to the function $F_k(\cdot\,;\vv_k^{t},\wg^{t})$. Clearly, $\nabla F_k(\cdot\,;\vv_k^{t},\wg^{t})$ is Lipschitz continuous with constant $L = (L_f+\sigma)$. Therefore, applying the above Lemma, we have the following after one gradient descent iteration (starting with $\ww_k^{t}$) at the Local-Solve subroutine: $\forall k \in S_t,$
\BEA
F_k( \ww_k^{t+1}; \vv_k^{t},\wg^{t}) \le F_k( \ww_k^{t}; \vv_k^{t},\wg^{t})  - \frac{ \| \nabla F_k(\ww_k^{t}; \vv_k^{t},\wg^{t}) \|_2^2}{2 L}. \nonumber
\EEA
Now, note the fact that $F_k( \ww_k^{t+1}; \vv_k^{t},\wg^{t})$ remains non-increasing after each gradient descent step. This completes the proof.

%%%%----------------------------------------------------------

\subsection{Proof of Theorem~\ref{thm:thm_new_prob_det}}

We start the proof with following observations from \eqref{eq:aggregation_def} and \eqref{eq:zk_rp_def}:
\BEA
\wg^{t} ~&=&~ \argmin_{ \wg \in \RR^d } \Big[\, \min_{\V \in \RR^{d \times \K}} H_{\sigma}(\W^t,\V,\wg)\, \Big] ,\nonumber \\
(\vv_1^{t}, \ldots, \vv_\K^t)~&=&~ \argmin_{ \V \in \RR^{d \times \K}  } H_{\sigma}(\W^t,\V,\wg^t),~t=1,2,\ldots \nonumber
\EEA
Combining the above, we have the following: $\forall t \ge 1,$
\BEA
(\vv_1^{t}, \ldots, \vv_\K^t,\wg^{t}) ~=~ \argmin_{ \V \in \RR^{d \times \K}, \,\wg \in \RR^d } H_{\sigma}(\W^t,\V,\wg). \label{eq:Z_w_updatex} 
\EEA
This implies
\BEA	
H_{\sigma}(\W^t,\V^t,\wg^{t}) ~\le ~ H_{\sigma}(\W^t,\V^{t-1},\wg^{t-1}), ~t=1,2,\ldots \label{eq:H_Zw_decrease}
\EEA
Before moving further, we introduce the following notation $F_k^t(\ww) := F_k(\ww; \vv_k^{t},\wg^{t})$. Now, we have the following from Proposition~\ref{prop:grad_decrease}:
\BEA
F_k^t( \ww_k^{t+1}) ~\le~ F_k^t( \ww_k^{t})  - \frac{1}{2 L} \| \nabla F_k^t(\ww_k^{t}) \|_2^2,\,\forall k \in \S^t \label{eq:Fkdecrease1}
\EEA
where $L := (L_f+\sigma)$. Moreover, $\ww_k^{t+1} = \ww_k^{t}$ for all $k \notin \S^t$ implies that
\BEA
F_k^t( \ww_k^{t+1}) ~\le~ F_k^t( \ww_k^{t}), ~\forall  k \notin \S^t. \label{eq:Fkdecrease2}
\EEA
Summing \eqref{eq:Fkdecrease1} and \eqref{eq:Fkdecrease2}, we get: $\forall t=0,1,\ldots,$
\BEA	
H_{\sigma}(\W^{t+1},\V^{t},\wg^{t}) ~\le ~ H_{\sigma}(\W^{t},\V^{t},\wg^{t}) - \frac{1}{2 \K L} \sum_{k \in \S^t }\| \nabla F_k^t(\ww_k^{t}) \|_2^2,  \label{eq:H_X_decrease}
\EEA
We can also express \eqref{eq:H_X_decrease} in expectation form:
\BEA
\prb [ H_{\sigma}(\W^{t+1},\V^{t},\wg^{t})] ~\le ~ H_{\sigma}(\W^{t},\V^{t},\wg^{t})- \frac{p}{2 \K L} \sum_{k=1}^{\K} \| \nabla F_k^t(\ww_k^{t}) \|_2^2 ,  \label{eq:H_X_decrease_prb1}
\EEA
where the expectation is w.r.t the random subset $\S^t$ and $p \in(0,1]$ is the probability of $k \in \S^t$. Taking, expectations w.r.t  $\S^0,\S^1,\ldots,\S^t$ (i.e. all the randomness till round $t$), we get: $\forall t= 0,1,\ldots,$
\BEA
\prb [ H_{\sigma}(\W^{t+1},\V^{t},\wg^{t})] ~\le ~ \prb [ H_{\sigma}(\W^{t},\V^{t},\wg^{t})]  - \frac{p}{2 \K L} \sum_{k=1}^{\K} \prb [\| \nabla F_k^t(\ww_k^{t}) \|_2^2 ]. \label{eq:H_X_decrease_prb2}
\EEA
Combining \eqref{eq:H_X_decrease_prb2} and \eqref{eq:H_Zw_decrease}, we have: $\forall t= 0,1,\ldots,$
\BEA
\prb [ H_{\sigma}(\W^{t+1},\V^{t+1},\wg^{t+1})] ~\le~  \prb [ H_{\sigma}(\W^{t},\V^{t},\wg^{t})] - \frac{p}{2 \K L} \sum_{k=1}^{\K} \prb [\| \nabla F_k^t(\ww_k^{t}) \|_2^2 ]. ~ \label{eq:H_XZw_decrease_prb}
\EEA
Summing over all $t$ and using the fact $H_{\sigma}$ is bounded below, we arrive at \eqref{eq:exp_grad_zero}.

On the other hand, combining \eqref{eq:H_X_decrease} and \eqref{eq:H_Zw_decrease}, we get: $\forall t= 0,1,\ldots,$
\BEA
H_{\sigma}(\W^{t+1},\V^{t+1},\wg^{t+1}) ~\le ~  H_{\sigma}(\W^{t},\V^{t},\wg^{t})  - \frac{1}{2 \K L} \sum_{k \in \S^t} \| \nabla F_k^t(\ww_k^{t}) \|_2^2. ~ \label{eq:H_XZw_decrease}
\EEA
Now, from \eqref{eq:Z_w_updatex}  and \eqref{eq:Fmu}  we see that $F_\sigma(\W^t) = H_{\sigma}(\W^t,\V^{t},\wg^{t})$. Thus, from  \eqref{eq:H_XZw_decrease} we have that $\{ F_\sigma(\W^t) \}_{t=0}^\infty$ is monotonically non-decreasing; therefore, also converges to some real value say $\hat{F}_\mu$ because $H_{\sigma}$ is bounded below. 
The rest of the proof, when $f_k$s are convex, follows from Theorem 3.7 in \cite{Beck_alternate_minimization_convergence} as \eqref{eq:H_XZw_decrease} and \eqref{eq:Z_w_updatex} together suggest that Fed+ is basically an (approximate) alternating minimization approach for solving \eqref{eq:FPz}.

%%%%%%%%%%%%%%%%%%%%%%%%%%%%%%%%%%%%%%%%%%%%%%%%%%%%%%%%%%%%%%%%%%%%%%%%%%%%%%%%

\subsection{Proof of Theorem~\ref{thm:FixedPointThm}}
We start by introducing the following notation:
\BEA
\zz_k^t ~:= ~\wg^t + \vv_k^t, ~k=1,\ldots,\K, \, \forall t\ge 0.\label{eq:fixp3x}
\EEA
By assumption~\ref{ass:exact_prox_convex}, the Local-Solve subroutine in Fed+ returns $\ww_k^{t+1}$ as the exact minimizer of $F_k(\cdot\,;\vv_k^{t},\wg^{t})$, i.e., 
\BEA 
    \ww_k^{t+1} &:=& \argmin_{\ww_k} f_k(\ww_k) + \frac{\sigma}{2}\|\ww_k - (\vv_k^{t}+\wg^t)\|_2^2 \nonumber \\ &=& \prox_{f_k}^{\frac{1}{\sigma}}(\zz_k^{t}), ~\forall \,t \ge 0, \forall k . \label{eq:prox_fk}
\EEA
Now, we observe the following about Fed+: $\forall \, t\ge 0,$
\BEA
	\ww_k^{t+1} \! &=&\! \zz_k^{t} - \frac{1}{\sigma} \nabla  \hat{f}_k(\zz_k^{t}), ~k=1,\ldots,\K, \label{eq:fixp1}\\
	\wg^{t+1} \! &=&\! \frac{1}{\K} \sum_{k=1}^\K \ww_k^{t+1} \!-\! \frac{1}{\K} \sum_{k=1}^\K \prox_{\Psi}^{\frac{1}{\sigma}}(\ww_k^{t+1} - \wg^{t+1}), \label{eq:fixp2}\\
	\vv_k^t \! &=&\! \prox^{\frac{1}{\sigma}}_{\Psi} ( \ww_k^t - \wg^t), ~k=1,\ldots,\K, \label{eq:fixp3}
\EEA
where the 1st equation is a direct consequence of \eqref{eq:prox_fk}, the 2nd one comes from \eqref{eq:xhat_iter_phi} and the last one is by choice \eqref{eq:zk_rp_def}. Therefore, for a fixed point, the 2nd equation in \eqref{eq:FixPointEq} obviously hold. Now, for a fixed point, we also have the following from \eqref{eq:fixp2}: 
\BEA 	\wg^* ~&=& ~\frac{1}{\K} \sum_{k=1}^\K \ww_k^* \,-\, \frac{1}{\K} \sum_{k=1}^\K \prox_{\Psi}^{\frac{1}{\sigma}}(\ww_k^* - \wg^*). \label{eq:fixp4} \EEA
Replacing the first $\ww_k^*$ in \eqref{eq:fixp4} with $\zz_k^* - \frac{1}{\sigma} \nabla \hat{f}_k(\zz_k^*)$ and subsequently $\zz_k^*$ by $\wg^* + \prox^{\frac{1}{\sigma}}_{\Psi} ( \ww_k^* - \wg^*)$, we get
\BEA 	\wg^* ~&=&~ \frac{1}{\K} \sum_{k=1}^\K\Big [\zz_k^* - \frac{1}{\sigma} \nabla \hat{f}_k(\zz_k^*)  -  \prox_{\Psi}^{\frac{1}{\sigma}}(\ww_k^* - \wg^*) \Big]\nonumber \\
~&=&~ \frac{1}{\K} \sum_{k=1}^\K\Big [ \wg^*  - \frac{1}{\sigma}  \nabla \hat{f}_k(\zz_k^*) \Big] \nonumber  \\
 ~&=&~ \wg^* - \frac{1}{\sigma \K} \sum_{k=1}^\K \nabla \hat{f}_k(\zz_k^*).
 \nonumber
 \EEA
Thus, we have the first equation in \eqref{eq:FixPointEq}.

\subsection{Proof of Corollary~\ref{cor:fixpoint_coro}}
To prove (a), we apply Theorem~\ref{thm:FixedPointThm} with $\Psi = 0$. This choice of $\Psi$ leads to the choice $\vv_k = \ww_k-\wg$ from \eqref{eq:zk_rp_def}. Therefore, Fed+ boils to applying the proximal point algorithm $\ww_k^{t+1} = \prox_{f_k}^{\frac{1}{\sigma}}(\ww_k^{t}), ~t\ge 0,$ at each local party $k =1,\ldots,\K$. Therefore, we obtain the result \eqref{eq:fixa} as $\ww_k^* = \prox_{f_k}^{\frac{1}{\sigma}}(\ww_k^*)$ implies
\BEA
\ww_k^*=  \ww_k^* - \frac{1}{\sigma} \nabla \hat{f}_k(\ww_k^*) \implies  \ww_k^* \in \argmin_{\ww}f_k(\ww). \nonumber
\EEA

Next, we prove part (b) by applying Theorem~\ref{thm:FixedPointThm} with the following choice of $\Psi$: $\Psi(\ww) = 0$ iff $\ww = \zer$ and $+\infty$ otherwise. This particular $\Psi$ corresponds to the choice $\vv_k = \zer$ from \eqref{eq:zk_rp_def}. Also, the aggregation function $\A$ becomes the mean as from \eqref{eq:aggregation_def}. Now, putting $\zz_k^* = \wg^*$ in \eqref{eq:FixPointEq}, we arrive at \eqref{eq:fixb}.

Finally, we show part (c) by setting $\Psi(\ww) = \frac{\sigma \delta}{2}\|\ww\|_2^2,\,\ww\in \RR^d$ in Theorem~\ref{thm:FixedPointThm}. In this case, \eqref{eq:zk_rp_def} becomes $\vv_k = [1+\delta]^{-1} ( \ww_k -\wg)$. Also, like in part (b), the aggregation function $\A$ becomes the mean here as well. Now, we complete the proof by using $\zz_k^* = \frac{\ww_k^* + \delta \wg^*}{1+\delta}$ in \eqref{eq:FixPointEq}:
\BEA
	\ww_k^* = \frac{\ww_k^* + \delta \wg^*}{1+\delta}  - \frac{1}{\sigma} \nabla \hat{f}_k\big(\zz_k^*\big) \nonumber \\ \implies \ww_k^* ~=~ \wg^* - \left(\frac{1+\delta}{\sigma\delta} \right)\nabla \hat{f}_k\big(\zz_k^*\big). \nonumber
\EEA
%\end{proof}
Note that part (b) of the Corollary recovers the fixed point result of FedProx given in \citep{Pathak2020FedSplit}.

\end{document}